\documentclass[letterpaper, 10 pt, conference]{ieeeconf}

\IEEEoverridecommandlockouts
\usepackage[T1]{fontenc}
\usepackage[utf8]{inputenc}
\usepackage{cite}
\usepackage{graphicx}
\usepackage{booktabs}
\usepackage{amsmath}
\usepackage{amssymb}
\usepackage{array}
\usepackage{url}
\usepackage{placeins}
\usepackage{dblfloatfix}
\usepackage[hidelinks]{hyperref}

\graphicspath{{figures/}}

\title{\LARGE \bf
VIDAR: Visual-Inertial Dense Alignment and Reconstruction via a Geometric Foundation Model
}

\author{Diyari Mohammed Salih$^{1}$,
        Lingxiang Hu$^{1}$\thanks{Corresponding author: Lingxiang Hu (e-mail: hulxhlx@gmail.com).},
        Naima AitOufroukh-Mammar$^{1}$,
        and Fabien Bonardi$^{1}$%
\thanks{$^{1}$All authors are with IBISC Laboratory, Universit\'e d'\'Evry Paris-Saclay,
Universit\'e Paris-Saclay, \'Evry-Courcouronnes, France.}%
\thanks{E-mails: diyari.m.salih@gmail.com,
hulxhlx@gmail.com,
naima.aitoufroukh@univ-evry.fr,
fabien.bonardi@univ-evry.fr.}%
}

\begin{document}
\maketitle
\thispagestyle{empty}
\pagestyle{empty}

\begin{abstract}
Monocular foundation models provide dense geometry but usually lack a stable metric scale. This paper presents VIDAR, a visual-inertial dense reconstruction framework that couples SVO+IMU odometry with Depth Anything 3. VIDAR uses the visual-inertial front end as a metric anchor: it provides camera poses, scale, and a consistent world frame for aligning dense foundation-model predictions across time. The foundation model then contributes detailed local geometry that is fused into a global reconstruction. We study both pose-conditioned DA3 and a decoupled alignment strategy. On EuRoC, pose injection reduces scale error to about 1\% and reaches 0.463 mean F@0.10; the decoupled hybrid improves this to 0.676 without ground-truth poses. Results on EuRoC and TUM RGB-D show that VIDAR is a practical route to metric dense monocular reconstruction.
\end{abstract}

% Keywords: visual-inertial odometry, dense reconstruction, geometric foundation
% model, Depth Anything 3, SVO, monocular mapping, robotics perception.

\section{Introduction}

Mobile robots require dense geometric maps for inspection, navigation, obstacle reasoning, and downstream scene understanding. Related dense reconstruction and robotic perception systems also combine feed-forward 3D reconstruction, sparse-view refinement, navigation, detection, segmentation, and task-specific deployment constraints for real-time operation~\cite{hu2025ec3rslam,hu2025dprsplat,hu2024hybridnavigation,hu2024handkeypoint,zhu2025metaldefect,li2025multiuav}. At the same time, practical onboard systems must estimate camera motion reliably under limited compute, memory, and sensing conditions. Classical visual or visual-inertial odometry (VO/VIO) provides an efficient and interpretable motion backbone, but it does not by itself provide dense scene geometry. Recent geometric foundation models, such as Depth Anything 3 (DA3), provide a complementary capability: they infer dense monocular geometry and can also produce camera-pose estimates from image sequences. However, when used as monocular systems, their reconstructions remain tied to a learned and often unstable native scale.

\begin{figure*}[!t]
\centering
\includegraphics[width=0.98\textwidth]{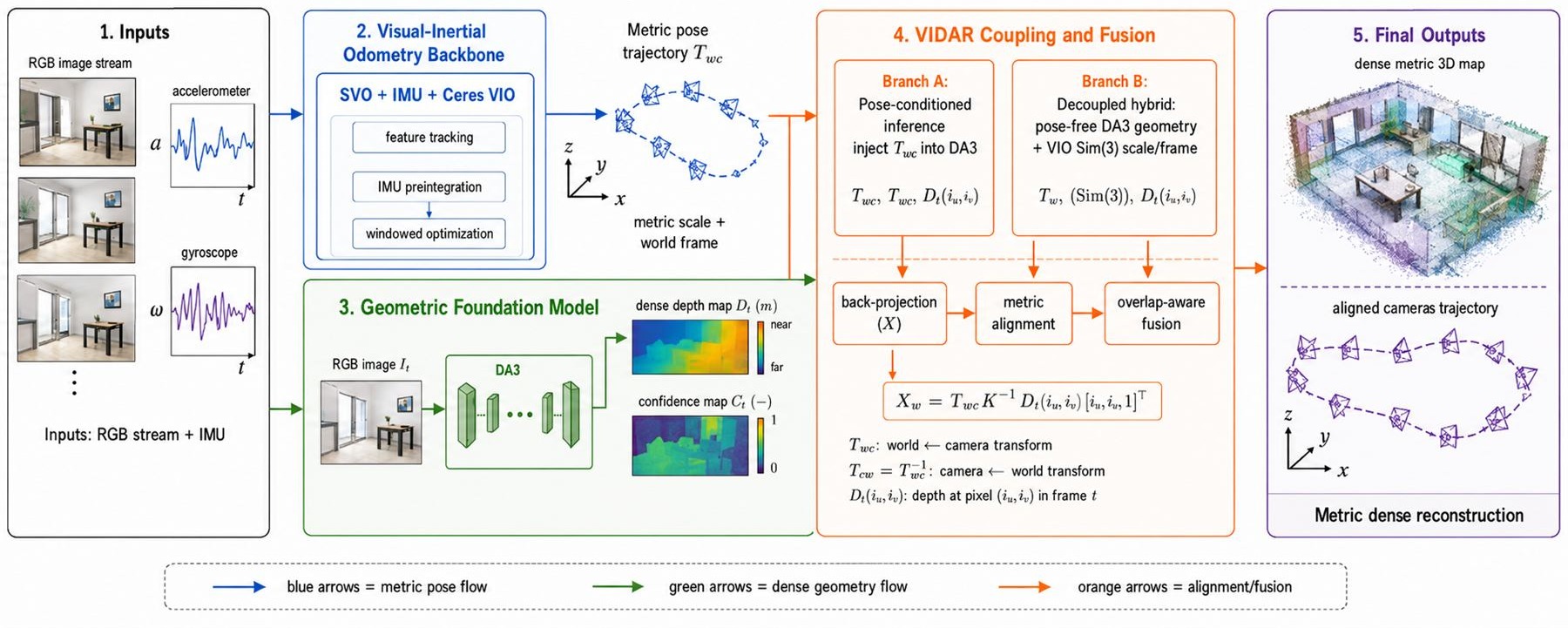}
\caption{Overview of the proposed VIDAR pipeline. SVO+IMU estimates metric poses, while DA3 predicts dense geometry that is conditioned on, or subsequently aligned to, the VIO frame.}
\label{fig:vidar_system_overview}
\end{figure*}

This paper argues that these two capabilities should be coupled rather than treated as mutually exclusive alternatives. A geometric foundation model can provide dense local geometry that is difficult to obtain from lightweight VO alone. Conversely, SVO+IMU can provide efficient metric motion when synchronized inertial measurements are available. The missing link is metric dense mapping: the foundation model needs scale and a stable world frame, while the VIO front end needs dense geometry. These observations motivate a hybrid framework in which visual-inertial odometry supplies metric scale and frame alignment, and the foundation model supplies dense local structure.

We propose VIDAR, a Visual-Inertial Dense Alignment and Reconstruction framework via a geometric foundation model. VIDAR uses SVO+IMU as a lightweight VIO front end and DA3/DA3-Streaming as the dense monocular geometry module. The central design principle is that dense reconstruction should inherit metric scale and global placement from the VIO trajectory, while the foundation model should preserve its strong local geometric prior.

The paper evaluates this design using EuRoC MAV and TUM RGB-D. EuRoC provides a visual-inertial setting in which SVO+IMU can be tested as the metric anchor and DA3 reconstructions can be evaluated against Leica dense reference scans. TUM RGB-D provides an RGB-only setting in which learned monocular priors can be compared against monocular SVO. The objective is not to claim that either module dominates universally, but to establish the complementary roles required by VIDAR: robust visual-inertial metric anchoring and dense foundation-model geometry.

The contributions of this paper are as follows:

\begin{itemize}
\item We propose VIDAR, a hybrid dense reconstruction framework that combines lightweight SVO+IMU odometry with a geometric foundation model for monocular dense geometry.
\item We formulate two VIO-coupling strategies for DA3: pose-conditioned inference using injected VIO extrinsics, and a decoupled hybrid that preserves pose-free DA3 geometry while applying VIO-derived scale and frame afterward.
\item We evaluate trajectory anchoring and dense reconstruction behavior across EuRoC and TUM RGB-D sequences.
\item We show that VIO resolves the metric-scale ambiguity of pose-free DA3 on EuRoC, while the decoupled hybrid gives the strongest dense reconstruction result without using ground truth.
\end{itemize}

All quantitative claims are tied to the reported experiments. The proposed framework is therefore presented as a practical hybrid reconstruction direction supported by trajectory, reconstruction, and resource evidence.

\section{Related Work}

\subsection{Visual-Inertial Odometry as a Geometric Anchor}

Visual and visual-inertial odometry estimate the camera trajectory from image measurements, and in the inertial case from synchronized IMU measurements. SVO estimates motion with a semi-direct formulation combining sparse image alignment, feature alignment, pose refinement, and probabilistic depth filtering~\cite{forster2017svo}. Its efficiency makes it suitable as a lightweight front end for resource-constrained robots. In the VIDAR framework, SVO+IMU is used not as a dense mapper, but as a metric geometric anchor that can constrain the coordinate frame in which dense reconstruction is accumulated.

Other classical systems provide complementary design points. Feature-based SLAM systems such as ORB-SLAM and ORB-SLAM3~\cite{mur2015orbslam,campos2021orbslam3} emphasize keypoint matching, loop closure, and global consistency. DSO~\cite{engel2018dso} follows a direct sparse formulation based on photometric consistency, while RTAB-Map~\cite{labbe2019rtabmap} extends toward large-scale RGB-D mapping. These methods motivate the broader SLAM context, but this work focuses on SVO because it provides a lightweight geometric backbone with a mono-inertial mode suitable for the proposed hybrid pipeline.

\subsection{Geometric Foundation Models}

This paper uses the term geometric foundation model to describe large-scale pretrained feedforward models that infer scene-level geometric quantities from visual observations, such as monocular depth, dense structure, camera motion, or local 3D reconstruction. DA3 is used here as a representative geometric foundation model~\cite{lin2025depthanything3}. Its value for VIDAR is not only trajectory estimation, but also dense geometry generation from RGB input. DA3-Streaming is especially relevant for long sequences because it carries temporal state while estimating both geometry and camera motion.

Related learned reconstruction systems, including EC3R-SLAM~\cite{hu2025ec3rslam}, DPR-Splat~\cite{hu2025dprsplat}, VGGT~\cite{vggt}, MapAnything~\cite{mapanything}, and Fast3R~\cite{yang2025fast3r}, illustrate the growing space of feedforward geometric reconstruction and sparse-view 3D refinement. They are treated as related directions rather than full quantitative baselines in this paper. The focus is instead on how a representative geometric foundation model can be coupled with visual-inertial odometry for dense robotic reconstruction.

\subsection{Dense Alignment and Mapping}

Dense mapping requires more than estimating a plausible local depth map. Local reconstructions must be placed into a consistent global frame, and long sequences must avoid scale drift, pose drift, and local geometric inconsistency. Monocular trajectory evaluation commonly uses Sim(3) alignment because metric scale may be unknown; accordingly, this paper reports Sim(3)-aligned RMSE as the common trajectory-shape metric. However, Sim(3) alignment is only a global similarity transform. It cannot repair local non-rigid differences between independently reconstructed learned chunks. This limitation is central to the VIDAR design: visual-inertial odometry can provide the metric anchor, but the dense foundation-model reconstruction must also preserve temporal consistency.

\section{Proposed Method: VIDAR}

\subsection{System Overview}

Given an RGB image sequence \(\{I_i\}\) and, when available, synchronized IMU measurements, VIDAR aims to reconstruct a dense scene in a metric world frame. This differs from pose-free feed-forward reconstruction, where geometry is often recovered in an arbitrary learned scale and coordinate frame. VIDAR explicitly separates the problem into two coupled components: metric frame estimation and dense local geometry prediction. The SVO/SVO+IMU front end estimates camera motion in a physically meaningful scale, while DA3 predicts dense monocular geometry from visual input. The final reconstruction inherits local surface detail from the geometric foundation model and metric placement from the visual-inertial trajectory.

Figure~\ref{fig:vidar_system_overview} summarizes the proposed visual-inertial anchoring, dense geometry prediction, and fusion pipeline.

The method is motivated by a division of labor. SVO+IMU is efficient and metric when inertial data are available, making it suitable for scale and frame anchoring. DA3 provides dense local geometry, making it suitable for reconstruction. VIDAR therefore treats visual-inertial odometry as the metric backbone and DA3 as the dense geometry module.

\subsection{Visual-Inertial Geometric Anchoring}

Let \(T_{wi}\) denote the pose of camera frame \(i\) in the world frame estimated by the visual-inertial front end. In VIDAR, this pose is used as a metric anchor for dense reconstruction. Rather than accepting learned camera poses blindly, the system uses the visual-inertial trajectory to define the coordinate frame into which dense predictions are inserted.

For a pixel \((u,v)\) with predicted depth \(D_i(u,v)\) and camera intrinsics \(K\), the corresponding local 3D point is
\begin{equation}
\mathbf{x}_i = D_i(u,v) K^{-1} [u, v, 1]^T .
\end{equation}
The point is then placed in the world frame using the anchored camera pose,
\begin{equation}
\mathbf{x}_w = T_{wi}\mathbf{x}_i .
\end{equation}
This formulation captures the core VIDAR principle: dense geometry is produced by the foundation model, while metric scale and frame are provided by visual-inertial odometry.

\subsection{Geometric Foundation Model Reconstruction}

DA3 is used as the representative geometric foundation model. It provides dense monocular geometry and, in DA3-Streaming mode, learned long-sequence pose estimates. In VIDAR, these learned outputs are interpreted according to their role. Dense geometry is used as the primary reconstruction signal. Learned poses are useful in RGB-only domains and as a comparison point, but they do not remove the need for a visual-inertial anchor when synchronized IMU data are available and metric stability is required.

\subsection{Pose-Conditioned and Decoupled Coupling}

VIDAR evaluates two ways to couple DA3 with the VIO trajectory. In the pose-conditioned variant, the DA3-Streaming inference call receives external camera extrinsics derived from SVO+IMU poses. If \(T_{WB}\) is the body-to-world pose and \(T_{BC}\) is the camera extrinsic, the camera-to-world pose is \(T_{WC}=T_{WB}T_{BC}\), and the injected world-to-camera pose is
\begin{equation}
T_{CW}=T_{WC}^{-1}=(T_{WB}T_{BC})^{-1}.
\end{equation}
The model is then called with external poses and intrinsics, \(\mathrm{DA3}(I;\mathrm{ext}=T_{CW},\mathrm{intr}=K)\). To make each chunk metric, VIDAR estimates a robust Umeyama scale \(s^\star\) between DA3-predicted camera centers and injected VIO centers and rescales the predicted depth,
\begin{equation}
s^\star=\operatorname{UmeyamaScale}(\{c_i^{\mathrm{pred}}\},\{c_i^{\mathrm{VIO}}\}),\qquad
D\leftarrow D/s^\star .
\end{equation}
Thus DA3 predicts dense geometry while being conditioned on a metric pose sequence.

The second variant is a decoupled hybrid. DA3 first reconstructs in its pose-free native frame, preserving its local geometric prior. A single similarity transform is then estimated between the DA3-predicted camera centers and the VIO camera centers,
\begin{equation}
S_{\mathrm{vio}} = \operatorname{Umeyama}(\{c_i^{\mathrm{DA3}}\}, \{c_i^{\mathrm{VIO}}\}),
\end{equation}
and the dense map is transformed as \(M_{\mathrm{VIDAR}}=S_{\mathrm{vio}}(M_{\mathrm{DA3}})\). This keeps DA3's local geometry while using VIO only for metric scale and global frame.

\begin{figure*}[!t]
\centering
\includegraphics[width=0.98\textwidth]{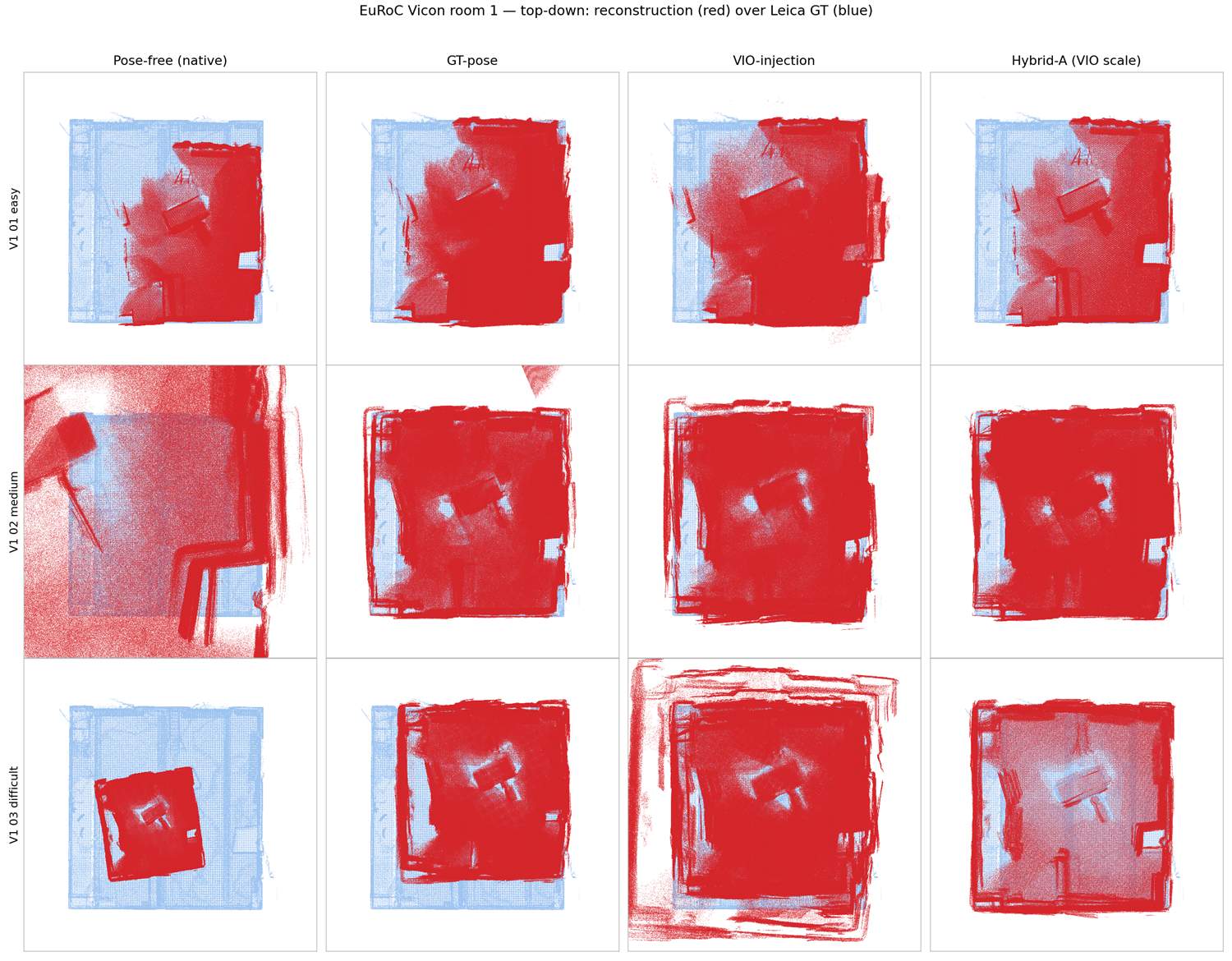}
\caption{EuRoC V1 dense reconstruction comparison against the Leica scan. Pose-free DA3 can recover local structure but suffers from native-scale errors, while VIO-based alignment keeps the reconstruction metric; the decoupled hybrid preserves DA3 geometry and uses VIO scale and frame.}
\label{fig:euroc_v1_dense_reconstruction}
\end{figure*}

\subsection{Dense Alignment and Fusion}

The dense alignment problem is to transform DA3-generated local geometry into a consistent scene map. The straightforward version of VIDAR uses the SVO+IMU trajectory to align frame-level or chunk-level dense geometry. Adjacent local reconstructions are connected using overlap-based Sim(3) alignment when overlap is sufficiently rigid and geometrically consistent. Given overlapping source and target point sets \(P_s,P_t\subset\mathbb{R}^3\), with means \(\mu_s,\mu_t\) and RMS radii \(\sigma_s,\sigma_t\), VIDAR uses the closed-form Umeyama similarity
\begin{align}
s &= \sigma_t/\sigma_s,\qquad
H=(s(P_s-\mu_s))^T(P_t-\mu_t),\\
U\Sigma V^T&=\operatorname{SVD}(H),\quad
R=V\operatorname{diag}(1,1,\det(VU^T))U^T,\\
t&=\mu_t-sR\mu_s .
\end{align}
A Sim(3) element \(S=(s,R,t)\) acts on points as \(S(p)=sRp+t\), and chunk poses are accumulated as \(S_k=S_{k-1}\circ\Delta S_k\).

\subsection{Loop Closure and Robustness Guards}

For streaming reconstruction, VIDAR follows DA3-Streaming in using image-based loop constraints over chunks. A global place-recognition descriptor is computed for each chunk reference image; a loop edge is declared between chunks \(i,j\) when the descriptor cosine similarity exceeds \(\tau_{\mathrm{loop}}\). The relative constraint \(\Delta S_{ij}\) is obtained with the same point-map Umeyama alignment, and a Sim(3) pose graph refines the absolute chunk poses:
\begin{equation}
\min_{\{S_k\}}\sum_{(i,j)\in\mathcal{E}}
\left\|\operatorname{Log}_{\mathrm{Sim}(3)}(\Delta S_{ij}^{-1}S_i^{-1}S_j)\right\|^2 .
\end{equation}
Here \(\mathcal{E}\) contains sequential and loop edges. The SVO/VIO backbone is run without its own loop closure; loop closure is applied only inside the DA3-Streaming chunk graph.

Injected metric poses also make abnormal scale estimates easy to detect. VIDAR therefore clamps invalid inter-chunk scales and guards degenerate Umeyama fits:
\begin{align}
\hat{s}=
\begin{cases}
s, & s\in[0.85,1.20],\\
1, & \mathrm{otherwise},
\end{cases}\\
\mathrm{rank\ deficient}&\Rightarrow (s,R,t)=(1,I,0).
\end{align}
These guards prevent individual failed chunks from corrupting the cumulative Sim(3) chain.

Learned monocular chunks may still differ by local shape deformation, so a single Sim(3) can be insufficient for long independent chunk fusion. The decoupled hybrid is best suited to short or low-drift sequences, while longer sequences motivate a coupled pose graph in which DA3 chunks retain their internal geometry and VIO relative constraints place them metrically.

\begin{figure*}[!t]
\centering
\includegraphics[width=0.82\textwidth]{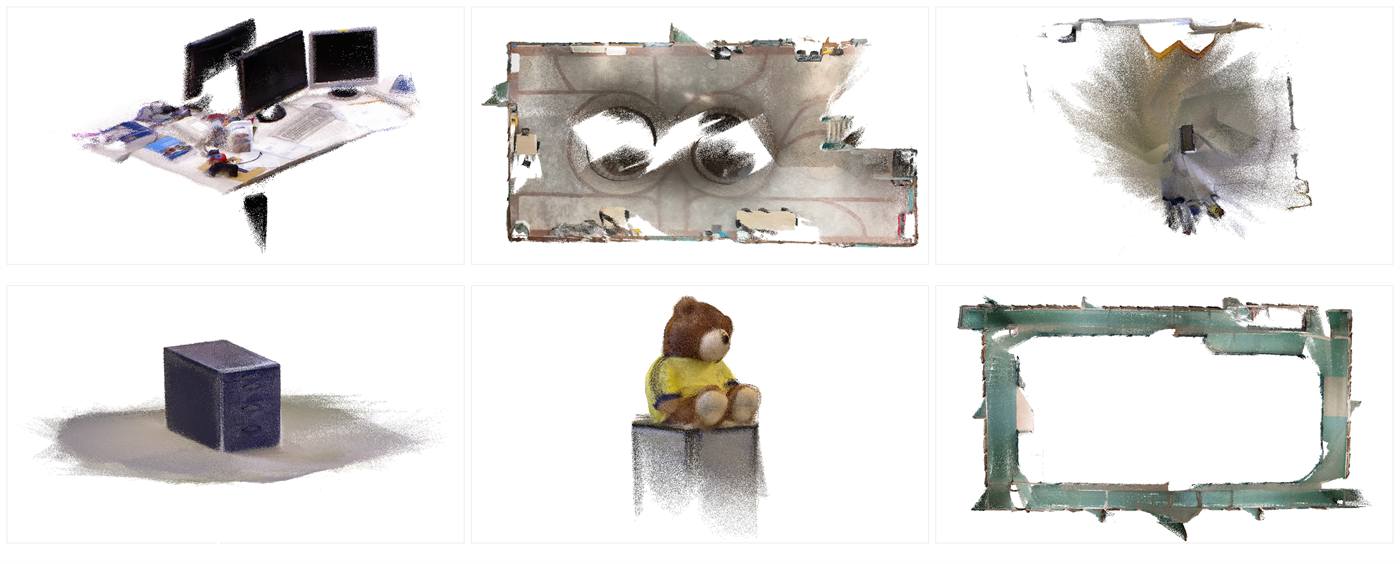}
\caption{Qualitative DA3 reconstruction examples from Freiburg scenes.}
\label{fig:da3_reconstruction_montage}
\end{figure*}

\section{Experiments}

\subsection{Datasets}

The experiments use EuRoC MAV~\cite{burri2016euroc} and TUM RGB-D~\cite{sturm2012tumrgbd}. EuRoC provides the main visual-inertial setting over 11 standard sequences, allowing SVO+IMU to be tested as a metric trajectory anchor. TUM RGB-D is used as an RGB-only trajectory and reconstruction setting; depth is not used as input to either SVO or DA3.

The quantitative evaluation focuses on the two modules used by VIDAR: SVO/SVO+IMU as the lightweight geometric anchor and DA3/DA3-Streaming as the geometric foundation-model component. On EuRoC, dense reconstructions are also compared with the Leica point cloud reference. This allows the study to separate trajectory accuracy from metric dense-map quality.

\subsection{Evaluation Protocol and Metrics}

The experiments evaluate the components required by VIDAR. SVO and SVO+IMU are evaluated as trajectory anchors, while DA3 and DA3-Streaming are evaluated as geometric foundation-model components that provide learned poses and dense reconstruction. Trajectory quality is measured by Sim(3)-aligned RMSE, which measures trajectory shape after similarity alignment and should not be interpreted as raw metric-scale accuracy.

Dense EuRoC reconstructions are evaluated against Leica scans using F-score and scale error, exposing whether a reconstruction is metric before any ground-truth scale is borrowed. The dense reference is the Leica scan and the reported dense metric is F-score after native-scale bounded Sim(3) placement. The cloud is first placed at its native scale by an SE(3) fit of its pose-source trajectory to GT, followed by a single scale refine accepted only in \([0.85,1.15]\):
\begin{equation}
T^\star=\arg\min_{\substack{T\in\mathrm{Sim}(3)\\s(T)\in[0.85,1.15]}}
\sum_x \min_y \|Tx-y\|^2 .
\end{equation}
This corrects small uniform depth inflation but prevents a pose-free monocular method from freely borrowing arbitrary ground-truth scale. For \(\tau\in\{0.05,0.10\}\) m,
\begin{align}
P&=\frac{|\{x:\min_y\|x-y\|<\tau\}|}{|\mathcal{R}|},\\
R&=\frac{|\{y:\min_x\|x-y\|<\tau\}|}{|\mathcal{G}|},\\
F@\tau&=\frac{2PR}{P+R}.
\end{align}
Metric-ness is summarized by the trajectory scale error \(|\ln s|\), where \(s\) is the Umeyama scale to the ground-truth trajectory. Runtime and memory are measured on a benchmark machine with approximately 31~GiB of system RAM and an NVIDIA RTX 4070-class GPU with approximately 12~GiB of VRAM; exploratory checks from an RTX 2060 6~GB laptop are used only as feasibility evidence.

\begin{table*}[!t]
\centering
\caption{Sim(3)-aligned trajectory RMSE summary. Values are averaged over valid sequence-level estimates. For TUM RGB-D, monocular SVO failed on one sequence, so the SVO aggregate uses the five valid SVO outputs.}
\label{tab:trajectory_metrics}
\small
\begin{tabular}{llccc}
\toprule
Dataset & Method & Valid sequences & Mean RMSE (m) & Median RMSE (m) \\
\midrule
EuRoC MAV & SVO mono & 11/11 & 0.898 & 0.740 \\
EuRoC MAV & SVO mono-IMU & 11/11 & 0.125 & 0.120 \\
EuRoC MAV & DA3-Streaming, 5 fps & 11/11 & 2.228 & 1.710 \\
TUM RGB-D & SVO mono & 5/6 & 0.429 & 0.369 \\
TUM RGB-D & Best DA3 per sequence & 6/6 & 0.031 & 0.023 \\
\bottomrule
\end{tabular}
\end{table*}

\subsection{Trajectory Backbone Validation}

The first experiment evaluates whether SVO+IMU is a suitable metric backbone for VIDAR. Table~\ref{tab:trajectory_metrics} summarizes the trajectory results. On EuRoC, adding inertial constraints reduces the mean Sim(3)-aligned RMSE from 0.898~m for monocular SVO to 0.125~m for mono-inertial SVO, with the median decreasing from 0.740~m to 0.120~m. This supports using the VIO trajectory as the source of metric scale and frame.

The tested DA3-Streaming pose output is not sufficient to replace the VIO backbone in this visual-inertial setting. Its EuRoC mean RMSE is 2.228~m, higher than both monocular and mono-inertial SVO. These trajectory results motivate using DA3 primarily as the dense geometry source and VIO as the metric trajectory source.

\begin{figure*}[!t]
\centering
\begin{minipage}{0.43\textwidth}
\centering
\includegraphics[width=\linewidth]{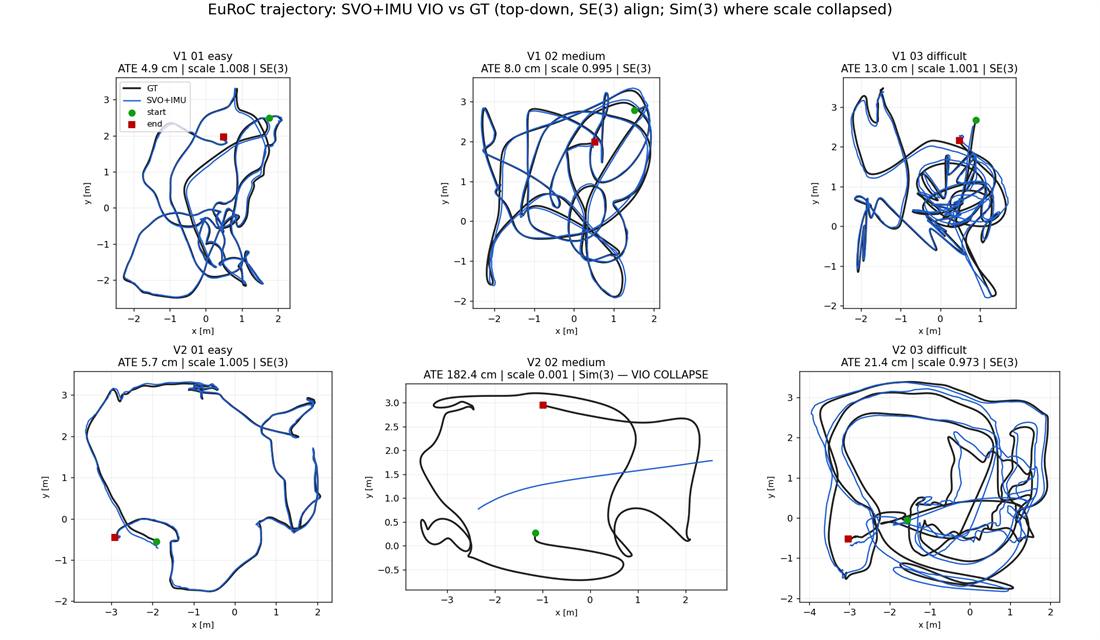}
\centerline{(a) SVO+IMU VIO vs. GT}
\end{minipage}
\hspace{0.04\textwidth}
\begin{minipage}{0.43\textwidth}
\centering
\includegraphics[width=\linewidth]{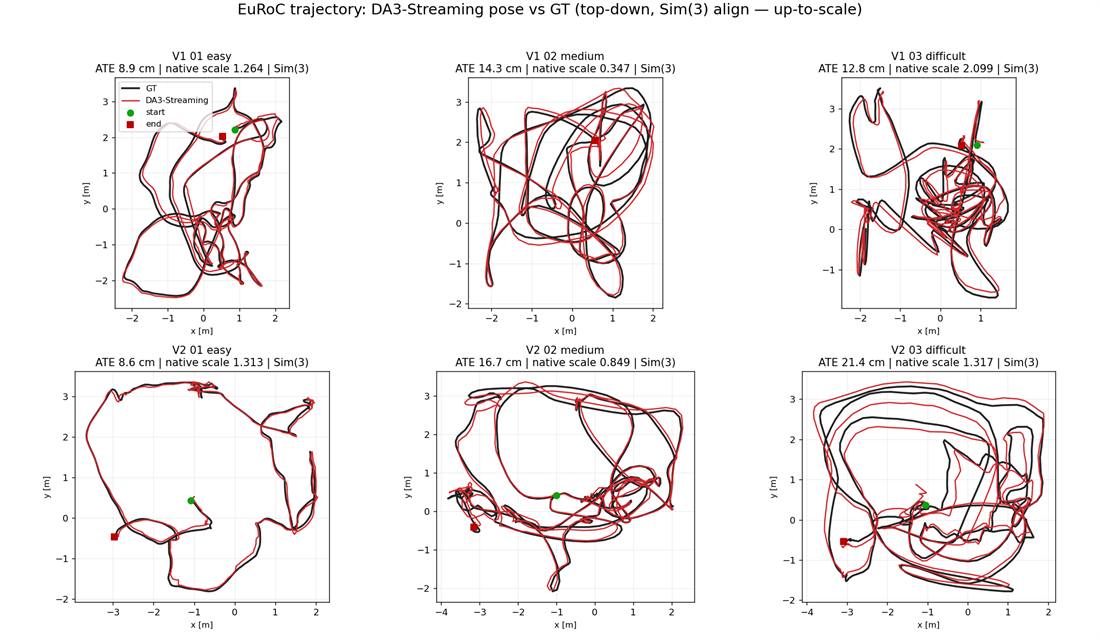}
\centerline{(b) DA3-Streaming pose vs. GT}
\end{minipage}
\caption{EuRoC top-down trajectory comparison.}
\label{fig:euroc_trajectory_svo_da3_comparison}
\end{figure*}

\subsection{Metric Dense Reconstruction}

\begin{table}[t]
\centering
\scriptsize
\setlength{\tabcolsep}{2pt}
\renewcommand{\arraystretch}{1.00}
\caption{EuRoC F-score against Leica GT. V2\_02 VIO is excluded from the VIO mean.}
\label{tab:vidar_dense_results}
\begin{tabular}{@{}lcccccc@{}}
\toprule
& \multicolumn{3}{c}{F@0.10} & \multicolumn{3}{c}{F@0.05} \\
\cmidrule(lr){2-4}\cmidrule(lr){5-7}
Sequence & PF & VIO & GT & PF & VIO & GT \\
\midrule
V1\_01 & 0.583 & 0.527 & 0.591 & 0.397 & 0.324 & 0.392 \\
V1\_02 & 0.025 & 0.528 & 0.618 & 0.013 & 0.322 & 0.437 \\
V1\_03 & 0.436 & 0.422 & 0.395 & 0.292 & 0.253 & 0.252 \\
V2\_01 & 0.554 & 0.545 & 0.539 & 0.394 & 0.338 & 0.422 \\
V2\_02 & 0.358 & -- & 0.586 & 0.215 & -- & 0.386 \\
V2\_03 & 0.476 & 0.291 & 0.295 & 0.299 & 0.169 & 0.197 \\
\midrule
\textbf{mean} & \textbf{0.405} & \textbf{0.463} & \textbf{0.504} & \textbf{0.268} & \textbf{0.281} & \textbf{0.348} \\
\bottomrule
\end{tabular}
\end{table}

\begin{table}[t]
\centering
\footnotesize
\setlength{\tabcolsep}{5pt}
\renewcommand{\arraystretch}{1.02}
\caption{Metric-ness summary. \(|\ln s|\) is scale error and \(\Delta F\) is the gain from borrowed GT scale.}
\label{tab:metricness}
\begin{tabular}{lccc}
\toprule
Metric & PF & VIO & GT \\
\midrule
mean \(|\ln s|\) & 0.458 & \textbf{0.009} & 0.000 \\
std\((\ln s)\) & 0.561 & \textbf{0.012} & 0.000 \\
F@0.10 native & 0.405 & 0.463 & 0.504 \\
F@0.10 free & 0.661 & 0.481 & \(\sim\)0.50 \\
\(\Delta F\) & \textbf{0.256} & 0.018 & \(\sim\)0 \\
\bottomrule
\end{tabular}
\end{table}

Table~\ref{tab:vidar_dense_results} reports the EuRoC dense reconstruction comparison against Leica reference scans, with representative reconstructions shown in Figure~\ref{fig:euroc_v1_dense_reconstruction}. Pose-free DA3 has strong local geometry but unstable native scale: Table~\ref{tab:metricness} shows mean scale error of 0.458, while VIO injection reduces it to 0.009. VIO injection reaches 0.463 mean F@0.10, close to the 0.504 GT-pose upper bound. This verifies that VIO supplies the metric information missing from monocular foundation-model reconstruction.

The decoupled hybrid is the strongest tested variant. It first reconstructs in the pose-free DA3 frame, then applies the VIO-derived Sim(3) scale and frame. By preserving DA3's local geometry and borrowing only the VIO metric placement, it increases mean F@0.10 to 0.676 without ground-truth poses. Thus, for short or low-drift sequences, the preferred VIDAR strategy is not hard pose replacement, but DA3 geometry with VIO scale and frame.

\subsection{Auxiliary RGB-Only Evaluation}

The RGB-only TUM RGB-D results provide auxiliary evidence that DA3 offers useful monocular geometric priors beyond EuRoC. On TUM RGB-D, monocular SVO has mean RMSE 0.429~m over five valid sequences, while the best tested DA3 runs average 0.031~m over six scenes. These results support DA3 as the dense geometry module when classical monocular tracking is weak.

\subsection{Runtime and Resource Cost}

Runtime and memory measurements highlight the deployment trade-off. SVO stays below 1~GB peak GPU memory, while DA3 uses 9560~MB on average and reaches 11125~MB. Figure~\ref{fig:da3_reconstruction_montage} shows why this cost may be worthwhile: VIDAR uses lightweight VIO for scale and frame, and DA3 when dense reconstruction justifies the GPU budget.

\section{Conclusion}

This paper proposed VIDAR, a visual-inertial dense alignment and reconstruction framework via a geometric foundation model. SVO+IMU provides the metric scale and world frame that monocular DA3 lacks, while DA3 supplies dense local geometry. On EuRoC, VIO injection reduces DA3 scale error to about 1\% and reaches 0.463 mean F@0.10; the decoupled hybrid improves to 0.676 without ground-truth poses. These results support VIDAR as a metric dense mapping direction in which VIO anchors scale and frame, and the foundation model provides dense reconstruction.

\begingroup
\scriptsize
\setlength{\parskip}{0pt}
\let\oldthebibliography\thebibliography
\let\endoldthebibliography\endthebibliography
\renewenvironment{thebibliography}[1]{%
  \oldthebibliography{#1}%
  \scriptsize%
  \setlength{\itemsep}{0pt}%
  \setlength{\parsep}{0pt}%
}{\endoldthebibliography}
\bibliographystyle{IEEEtran}
\bibliography{references}
\endgroup

\end{document}